%% file: main.tex
\documentclass[twoside,leqno,twocolumn]{article}
\pdfoutput=1 
\usepackage[letterpaper]{geometry}
\usepackage{ltexpprt}

\usepackage{cite}
\usepackage{amsmath,amssymb,amsfonts}
\usepackage{algorithmic}
\usepackage{graphicx}
\usepackage{textcomp}
\usepackage{grffile}
\usepackage{lipsum}
\usepackage{xcolor}
\usepackage[font=small]{caption}
\usepackage{subcaption}
\usepackage{wrapfig}
\usepackage[symbol]{footmisc}
\usepackage[export]{adjustbox}

\begin{document}

\newcommand{\nikhilc}[1]{\textcolor{red}{#1}}

\newcommand{\cfddem}{CFD-DEM}
\newcommand{\ourmethod}{PhyDNN\,}
\newcommand{\ourmethodAll}{PhyDNN\,}
\newcommand{\ourmethodPXTX}{PhyDNN-F$^P_x$F$^S_x$\,}

\title{\Large Physics-guided Design and Learning of Neural Networks \\ for Predicting Drag Force on Particle Suspensions in Moving Fluids}
  \author{Nikhil Muralidhar \footnotemark[1] \footnotemark[2]\\ \and
    Jie Bu \footnotemark[1] \footnotemark[2]\\ \and
    Ze Cao \footnotemark[3] \\ \and
    Long He \footnotemark[3]\\ \and 
    Naren Ramakrishnan \footnotemark[1]\\ \and
    Danesh Tafti \footnotemark[3]\\ \and 
    Anuj Karpatne \footnotemark[1] \footnotemark[2]\\ 
  }
 
\date{}

\maketitle

\footnotetext[1]{Dept.\ of Computer Science, Virginia Tech, VA, USA}
\footnotetext[2]{Discovery Analytics Center, Virginia Tech, VA, USA}
\footnotetext[3]{Dept.\ of Mechanical Engineering, Virginia Tech, VA, USA}

\begin{abstract} \small\baselineskip=9pt
Physics-based simulations are often used to model and understand complex physical systems and processes in domains like fluid dynamics. Such simulations, although used frequently, have many limitations which could arise either due to the inability to accurately model a physical process owing to incomplete knowledge about certain facets of the process or due to the underlying process being too complex to accurately encode into a simulation model. In such situations, it is often useful to rely on machine learning methods to fill in the gap by learning a model of the complex physical process directly from simulation data. However, as data generation through simulations is costly, we need to develop models, being cognizant of data paucity issues. In such scenarios it is often helpful if the rich physical knowledge of the application domain is incorporated in the architectural design of machine learning models. Further, we can also use information from physics-based simulations to guide the learning process using \textit{aggregate supervision} to favorably constrain the learning process. In this paper, we propose \ourmethod, a deep learning model using \textit{physics-guided structural priors} and \textit{physics-guided aggregate supervision} for modeling the drag forces acting on each particle in a \textit{Computational Fluid Dynamics-Discrete Element Method}(\cfddem). We conduct extensive experiments in the context of drag force prediction and showcase the usefulness of including physics knowledge in our deep learning formulation both in the \textit{design} and through \textit{learning} process. 
Our proposed \ourmethod model has been compared to several state-of-the-art models and achieves a significant performance improvement of \textbf{8.46}\% on average across all baseline models. The source code has been made available\footnote{https://github.com/nmuralid1/PhyDNN.git} and the dataset used is detailed in~\cite{he2017evaluation,he2019supervised}. 

\end{abstract}

\section{Introduction}
\input{sections/introduction.tex}
\section{Related Work}
\input{sections/related_work.tex}
\section{Proposed PhyDNN Framework}\label{sec:problem_formulation}
\input{sections/problem_formulation.tex}

\section{Dataset Description}\label{sec:dataset_description}
\input{sections/dataset_description.tex}
\section{Experimental Results}
\input{sections/results_and_discussion.tex}
\section{Conclusion}
\input{sections/conclusion.tex}

\bibliographystyle{ieeeTranFiles/IEEEtran}
\bibliography{bibfile}

\end{document}

%% file: sections/introduction.tex
Machine learning (ML) is ubiquitous in several disciplines today and with its growing reach, learning models are continuously exposed to new challenges and paradigms. In many applications, ML models are treated as black-boxes. In such contexts, the learning model is trained in a manner completely agnostic to the rich corpus of physical knowledge underlying the process being modeled. This domain-agnostic training might lead to many unintended consequences like the model learning spurious relationships between input variables, models learning representations that are not easily verifiable as being consistent with the accepted physical understanding of the process being modeled. Moreover, in many scientific disciplines, generating training data might be extremely costly due to the nature of the data generation collection process. To effectively be used across many scientific applications, it is important for data mining models to be able to leverage the rich physical knowledge in scientific disciplines to fill the void due to the lack of large datasets and be able to learn good process representations in the context of limited data. This makes the model less expensive to train as well as more interpretable due to the ability to verify whether the learned representation is consistent with the existing domain knowledge.
\begin{figure}
    \centering
    \includegraphics[scale=0.25]{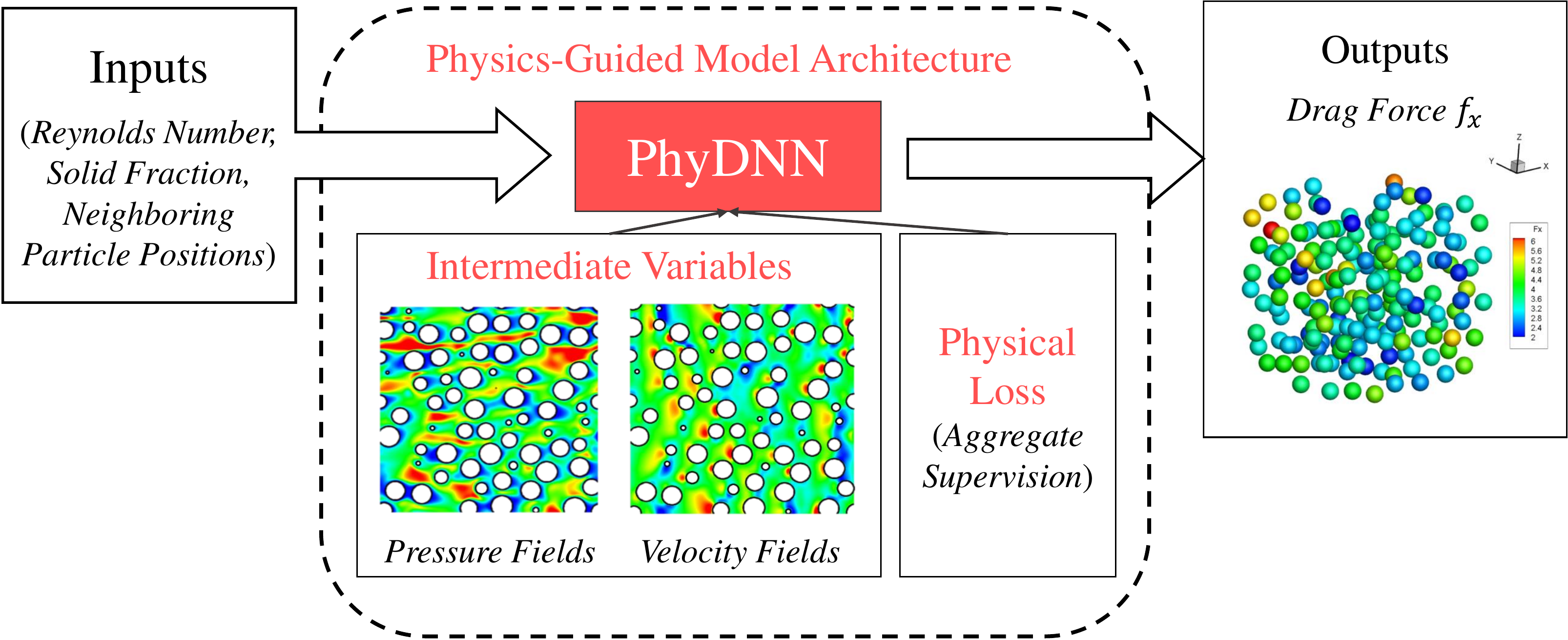}
    \caption{Our proposed PhyDNN Model.}
    \label{fig:my_label}
\end{figure}

In this paper, we attempt to bridge the gap between physics-based models and data mining models by incorporating domain knowledge in  the design and learning of machine learning models. Specifically, we propose three ways for incorporating domain knowledge in neural networks: (1) Physics-guided design of neural network architectures, (2) Learning with auxiliary tasks involving physical intermediate variables, and (3) Physics-guided aggregate supervision of neural network training.

We focus on modeling a system in the domain of multi-phase flows (solid particles suspended in moving fluid) which have a wide range of applicability in fundamental as well as industrial processes~\cite{li2003gas}. One of the critical interaction forces in these systems that has a large bearing on the dynamics of the system is the drag force applied by the fluid on the particles and vice-versa. The drag force can be obtained by Particle Resolved Simulations (PRS) at a high accuracy. It captures the velocity and pressure field surrounding each particle in the suspension that can later be used to compute the drag force. However, PRS is quite expensive and only a few 100s or at most 1000s of particles can be resolved in a calculation utilizing grids of O($10^8$) degrees of freedom and utilizing O($10^2$) processors or cores. Thus more practical simulations have to resort to coarse-graining, e.g., Discrete Element Method (DEM) and CFD-DEM. In these methods, a particle is treated as a point mass (not resolved) and the fluid drag force acting on the particles in suspension is modeled. 

Current practice in simulations is to use the mean drag force acting on the particle suspension as a function of flow parameters (Reynolds number) and the particle packing density (solid fraction - $\phi$) ~\cite{wen1966yh,di1994voidage,tenneti2011drag}. 
Given the variability of drag force on individual particles in suspension, this paper explores techniques in physics-guided machine learning to advance the current state-of-the-art for drag force prediction in CFD-DEM by learning from a small amount of PRS data. 

Our contributions are as follows:\\
{\noindent $\bullet$ We introduce \ourmethod, a novel \textit{physics-guided} model architecture that yields state-of-the-art results for the problem of particle drag force prediction.}\\  
{\noindent $\bullet$ We introduce \textit{physics-guided auxiliary tasks} to train \ourmethod more effectively with limited data.}\\
{\noindent $\bullet$ We augment \ourmethod architecture with \textit{aggregate supervision} applied over the auxiliary tasks to ensure consistency with physics knowledge. }\\
{\noindent $\bullet$ Finally, we conduct extensive experimentation to uncover several useful properties of our model in settings with limited data and showcase that \ourmethod is consistent with existing physics knowledge about factors influencing drag force over a particle, thus yielding greater model interpretability.}

%% file: sections/related_work.tex


There have been multiple efforts to leverage domain knowledge in the context of increasing the performance of data-driven or statistical models.  Methods have been designed to influence training algorithms in ML using domain knowledge, e.g., with the help of physically based priors in probabilistic frameworks \cite{wong2009active, xu2015robust,denli2014multi}, regularization terms in statistical models \cite{chatterjee2012sparse,liu2013accounting}, constraints in optimization methods \cite{majda2012physics, majda2012fundamental}, and rules in expert systems \cite{waterman1986guide,abu1990learning}. 
In a recent line of research, new types of deep learning models have been proposed (e.g., ODEnet \cite{chen2018neural} and RKnet \cite{zhu2018convolutional}) by treating sequential deep learning models such as residual networks and recurrent neural networks as discrete approximations of ordinary differential equations (ODEs). 

Yaser et al. show hints, i.e., prior knowledge can be incorporated into learning-from-example paradigm\cite{abu1990learning}. In~\cite{karpatne2017theory} the authors explored the idea of incorporating domain knowledge directly as a regularizer in neural networks to influence training and showed better generalization performance. In  \cite{Ren_Stewart_Song_Kuleshov_Ermon_2018, stewart2017label} domain knowledge was incorporated into a customized loss function for weak supervision that relies on no training labels.

There have been efforts to incorporate prior knowledge about a problem (like low rank structure of convolutional filters to be designed) into model architecture design (\textit{structural priors})~\cite{ioannou2018structural}. Also, to design neural network architectures, to incorporate feature invariance~\cite{ling2016reynolds}, implicit physics rules~\cite{seo2019differentiable} to enable learning representations consistent with physics laws and explicitly incorporating knowledge as constraints~\cite{leibo2017view}. In~\cite{anderson2019cormorant} the authors propose a neural network model where each individual neuron learns "laws" similar to physics laws applied to learn the behavior and properties of complex many-body physical systems. In~\cite{kondor2018generalization}, the authors propose a theory that details how to design neural network architectures for data with non-trivial symmetries. However none of these efforts are directly applicable to encode the physical relationships we are interested in modeling.

%% file: sections/problem_formulation.tex
\begin{figure*}[!ht]
    \centering
    \includegraphics[width=\textwidth]{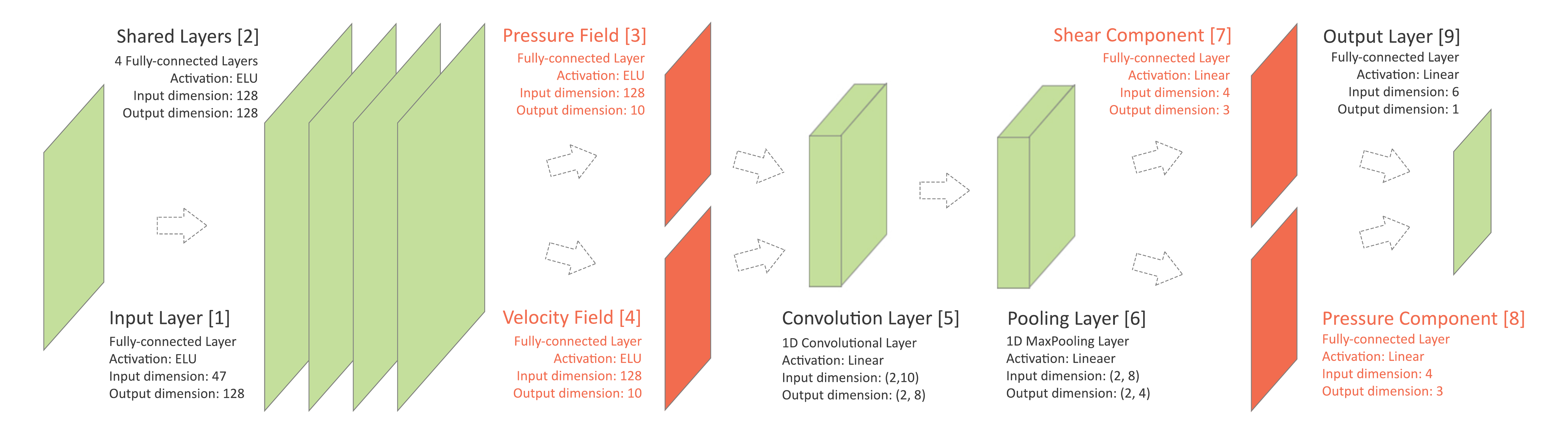}
    \caption{PhyDNN Architecture}
    \label{fig:model_architecture}
\end{figure*}

\subsection{Problem Background:}
Given a collection of $N$ 3D particles suspended in a fluid moving along the $X$ direction, we are interested in predicting the drag force experienced by the $i^\text{th}$ particle, $F_i$, along the $X$ direction due to the moving fluid. This can be treated as a supervised regression problem where the output variable is $F_i$, and the input variables include features capturing the spatial arrangement of particles neighboring particle $i$, as well as other attributes of the system such as Reynolds Number, Re,  and Solid Fraction (fraction of unit volume occupied by particles), $\phi$. Specifically, we consider the list of 3D coordinates of 15-nearest neighbors around particle $i$, appended with (Re, $\phi$) as the set of input features, represented as a flat 47-length vector, $\mathbf{A_i}$.

A simple way to learn the mapping from $\mathbf{A_i}$ to $F_i$ is by training feed-forward deep neural network (DNN) models, that can express highly non-linear relationships between inputs and outputs in terms of a hierarchy of complex features learned at the hidden layers of the network. However, black-box architectures of DNNs with arbitrary design considerations (e.g., layout of the hidden layers) can fail to learn generalizable patterns from data, especially when training size is small. To address the limitations of black-box models in our target application of drag force prediction, we present a novel physics-guided DNN model, termed \ourmethod{}, that uses physical knowledge in the design and learning of the neural network, as described in the following.

\subsection{Physics-guided Model Architecture:}
In order to design the architecture of PhyDNN, we derive inspiration from the known physical pathway from the input features $\mathbf{A_i}$ to drag force $F_i$, which is at the basis of physics-based model simulations such as Particle Resolved Simulations (PRS). Essentially, the drag force on a particle $i$ can be easily determined if we know two key physical intermediate variables: the pressure field ($\mathbf{P_i}$) and the velocity field ($\mathbf{V_i}$) around the surface of the particle. It is further known that $\mathbf{P_i}$ directly affects the \emph{pressure component} of the drag force, $F_i^P$, and $\mathbf{V_i}$ directly affects the \emph{shear component} of the drag force, $F_i^S$. Together, $F_i^P$ and $F_i^S$ add up to the total drag force that we want to estimate, i.e., $F_i = F_i^P + F_i^S$.

Using this physical knowledge, we design our PhyDNN model so as to express physically meaningful intermediate variables such as the pressure field, velocity field, pressure component, and shear component in the neural pathway from $\mathbf{A_i}$ to $F_i$. Figure \ref{fig:model_architecture} shows the complete architecture of our proposed PhyDNN model with details on the number of layers, choice of activation function, and input and output dimensions of every block of layers. In this architecture, the input layer passes on the 47-length feature vectors $\mathbf{A_i}$ to a collection of four \emph{Shared Layers} that produce a common set of hidden features to be used in subsequent branches of the neural network. These features are transmitted to two separate branches: the \emph{Pressure Field Layer} and the \emph{Velocity Field Layer}, that express $\mathbf{P_i}$  and $\mathbf{V_i}$, respectively, as 10-dimensional vectors. Note that $\mathbf{P_i}$ and $\mathbf{V_i}$ represent physically meaningful intermediate variables observed on a sequence of 10 equally spaced points on the surface of the particle along the $X$ direction. 

The outputs of pressure field and velocity field layers are combined and fed into a 1D Convolutional layer that extracts the sequential information contained in the 10-dimensional $\mathbf{P_i}$ and $\mathbf{V_i}$ vectors, followed by a Pooling layer to produce 4-dimensional hidden features. These features are then fed into two new branches, the \emph{Shear Component Layer} and the \emph{Pressure Component Layer}, expressing 3-dimensional $\mathbf{F_i^S}$ and $\mathbf{F_i^P}$, respectively. These physically meaningful intermediate variables are passed on into the final output layer that computes our target variable of interest: drag force along the $X$ direction, $F_i$. Note that we only make use of linear activation functions in all of the layers of our PhyDNN model following the Pressure Field and Velocity Field layers. This is because of the domain information that once we have extracted the pressure and velocity fields around the surface of the particle, computing $F_i$ is relatively straightforward. Hence, we have designed our PhyDNN model in such a way that most of the complexity in the relationship from $\mathbf{A_i}$ to $F_i$ is captured in the first few layers of the neural network. The layout of hidden layers and the connections among the layers in our PhyDNN model is thus physics-guided.
Further, the physics-guided design of PhyDNN ensures that we hinge some of the hidden layers of the network to express physically meaningful quantities rather than arbitrarily complex compositions of input features, thus adding to the interpretability of the hidden layers.

\subsection{Learning with Physical Intermediates:}
It is worth mentioning that all of the intermediate variables involved in our PhyDNN model, namely the pressure field $\mathbf{P_i}$, velocity field $\mathbf{V_i}$, pressure component $\mathbf{F_i^P}$, and shear component $\mathbf{F_i^S}$, are produced as by-products of the PRS simulations that we have access to during training. Hence, rather than simply learning on paired examples of inputs and outputs, $(\mathbf{A_i},F_i)$, we consider learning our PhyDNN model over a richer representation of training examples involving all intermediate variables along with inputs and outputs. Specifically, for a given input $\mathbf{A_i}$, we not only focus on accurately predicting the output variable $F_i$ at the output layer, but doing so while also accurately expressing every one of the intermediate variables  $(\mathbf{P_i},\mathbf{V_i},\mathbf{F_i^P},\mathbf{F_i^S})$ at their corresponding hidden layers. This can be achieved by minimizing the following empirical loss during training:
\begin{eqnarray}
\small 
  Loss_{MSE} = MSE(F,\widehat{{F}}) + \lambda_P ~ MSE(\mathbf{P},\widehat{\mathbf{P}})  \nonumber \\
   + \lambda_V ~ MSE(\mathbf{V},\widehat{\mathbf{V}}) 
  + \lambda_{FP} ~ MSE(\mathbf{F^P},\widehat{\mathbf{F^P}})  \nonumber \\ 
   + \lambda_{FS} ~ MSE(\mathbf{F^S},\widehat{\mathbf{F^S}})\nonumber 
\end{eqnarray}
where MSE represents the mean squared error, $\hat{x}$ represents the estimate of $x$, and $\lambda_P$, $\lambda_V$, $\lambda_{FP}$, and $\lambda_{FS}$ represent the trade-off parameters in miniming the errors on the intermediate variables. Minimizing the above equation will help in constraining our PhyDNN model with loss terms observed not only on the output layer but also on the hidden layers, grounding our neural network to a physically consistent (and hence, generalizable) solution. Note that this formulation can be viewed as a multi-task learning problem, where the prediction of the output variable can be considered as the primary task, and the prediction of intermediate variables can be viewed as auxiliary tasks that are related to the primary task through physics-informed connections, as captured in the design of our PhyDNN model.
 
\subsection{Using Physics-guided Loss:}
Along with learning our PhyDNN using the empirical loss observed on training samples, $Loss_{MSE}$, we also consider adding an additional loss term that captures our physical knowledge of the problem and ensures that the predictions of our PhyDNN model do not violate known physical constraints. In particular, we know that the distribution of pressure and velocity fields over different combinations of Reynolds number (Re) and solid fraction ($\phi$) show varying aggregate properties (e.g., different means), thus exhibiting a multi-modal distribution. If we train our PhyDNN model on data instances belonging to all (Re,$\phi$) combinations using $Loss_{MSE}$, we will observe that the trained model will under-perform on some of the modes of the distribution that are under-represented in the training set. To address this, we make use of a simple form of \emph{physics-guided aggregate supervision}, where we enforce 
the predictions $\mathbf{\widehat{P}_{(Re,\phi)}}$ and $\mathbf{\widehat{V}_{(Re,\phi)}}$ of the pressure and velocity fields around a particle respectively, at a given combination of (Re,$\phi$) to be close to the mean of the actual values of $\mathbf{P}$ and $\mathbf{V}$ produced by the PRS simulations at that combination. If $\overline{P}_{(Re,\phi)}$ and $\overline{V}_{(Re,\phi)}$ represent the mean of the pressure and velocity field respectively for the combination ($Re,\phi$), we consider minimizing the following physics-guided loss:
\begin{eqnarray}
\small 
  Loss_{PHY} = \sum_{Re} \sum_{\phi} MSE(\mu(\mathbf{\widehat{P}_{(Re,\phi)}}), \overline{P}_{(Re,\phi)})\nonumber \\
  + MSE(\mu(\mathbf{\widehat{V}_{(Re,\phi)}}), \overline{V}_{(Re,\phi)})\nonumber
\end{eqnarray}
The function $\mu(\cdot): R \xrightarrow{} R$ is a \emph{mean} function. 
We finally consider the combined loss $Loss_{MSE} + Loss_{PHY}$ for learning our PhyDNN model.

%% file: sections/dataset_description.tex
\iftrue
The dataset used has 5824 particles. Each particle has 47 input features including three-dimensional coordinates for fifteen nearest neighbors relative to the target particle's position, the Reynolds number ($Re$) and solid fraction ($\phi$) of the specific experimental setting (there are a total of 16 experimental settings with different ($Re$, $\phi$) combinations). Labels include the drag force in the X-direction $F_i \in \mathbb{R}^{1\times 1}$ as well as variables for auxiliary training, i.e., pressure fields ($\mathbf{P_i} \in \mathbb{R}^{10\times1}$), velocity fields ($\mathbf{V_i} \in \mathbb{R}^{10\times 1}$), pressure components ($\mathbf{F^P_i} \in \mathbb{R}^{3\times 1}$) and shear components of the drag force ($\mathbf{F^S_i} \in \mathbb{R}^{3\times 1}$)\footnote{Further details about the dataset included in the appendix.}.\par
\fi 
\subsection{Experimental Setup}\label{sec:experimental_setup}
All deep learning models used have 5 hidden layers, a hidden size of 128 and were trained for 500 epochs with a batch size of 100. Unless otherwise stated, 55\% of the dataset was used for training while the remaining data was used for testing and evaluation. We applied standardization to the all input features and labels in the data preprocessing step.

\textbf{Baselines:} We compare the performance of \ourmethod with several state-of-the-art regression baselines and a few close variants of \ourmethod.

{\noindent $\bullet$Linear Regression (Linear Reg.), Random Forest Regression (RF Reg.), Gradient Boosting Regression (GB Reg.)~\cite{scikitlearn}: We employed an ensemble of 100 estimators for RF, GB Reg. models and left all other parameters unchanged.}\\
{\noindent $\bullet$ DNN: A standard feed-forward neural network model for predicting the scalar valued particle drag force $F_i$.} \\  {\noindent $\bullet$ DNN+ Pres: A DNN model which predicts the pressure field around a particle ($\mathbf{P_i}$) in addition to $F_i$.}\\
{\noindent $\bullet$ DNN+ Vel: A DNN model which predicts the velocity field around a particle ($\mathbf{V_i}$) in addition to $F_i$.}\\
{\noindent $\bullet$ DNN-MT-Pres: Similar to DNN+ Pres except that the pressure and drag force tasks are modeled in a multi-task manner with a set of disjoint layers for each of the two tasks and a separate set of shared layers.}\\
{\noindent $\bullet$ DNN-MT-Vel; Similar to DNN-MT-Pres except in this case the auxiliary task models the velocity field around the particle ($\mathbf{V_i}$) in addition to drag force ($F_i$).}

\par\noindent We employ three metrics for model evaluation:


\textbf{MSE \& MRE}: We employ the mean squared error (MSE) and mean relative error (MRE)~\cite{he2019supervised} metrics to evaluate model performance. Though MSE can capture the absolute deviation of model prediction from the ground truth values, it can vary a lot for different scales of the label values, e.g., for higher drag force values, MSE is prone to be higher, vice versa. Thus, the need for a metric that is invariant to the scale of the label values brings in  the MRE as an important supplemental metric in addition to MSE.
\begin{equation*} 
\small
        MRE = \frac{1}{m}\sum_{i=1}^m \frac{|\widehat{F}_i - F_i|}{\overline{F}_{(Re,\phi)}}
\end{equation*}
$\overline{F}_{(Re,\phi)}$ is the mean drag force for $(Re,\phi)$ setting and $\widehat{F}_i$ the predicted drag force for particle i.
\begin{wrapfigure}{r}{0.26\textwidth}
    \includegraphics[scale=0.18,right]{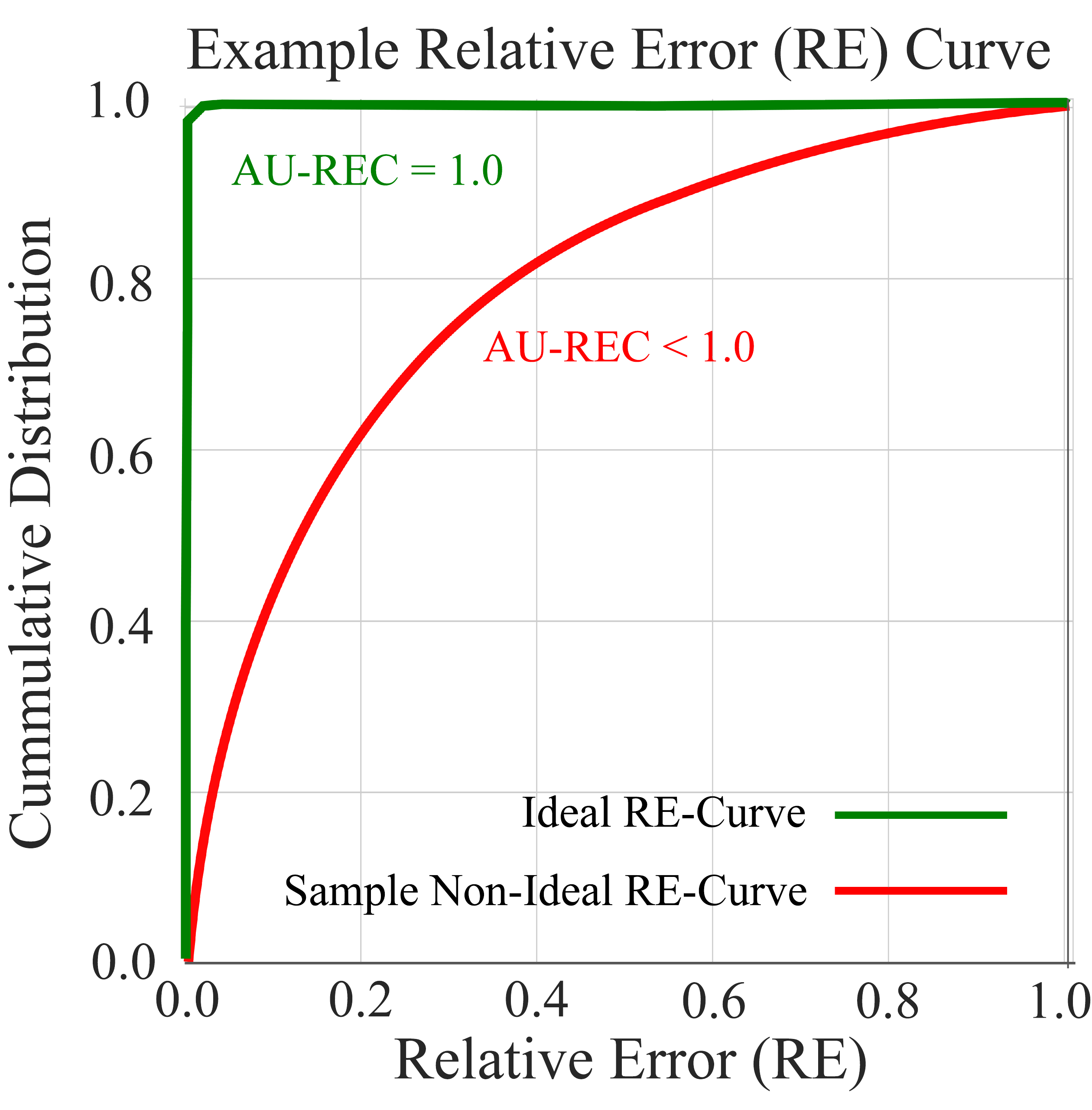}
    \label{fig:sample_rel_err_curve}
    \vspace{-0.5cm}
\end{wrapfigure}
\par\noindent\textbf{AU-REC}: The third metric we employ is the \textit{area under the relative error curve} (AU-REC). The relative error curve represents the cumulative distribution of relative error between the predicted drag force values and the ground truth PRS drag force data. AU-REC calculates the area under this curve. The AU-REC metric ranges between [0,1] and higher AU-REC values indicate superior performance.


%% file: sections/results_and_discussion.tex
We conducted multiple experiments to characterize and evaluate the model performance of \ourmethod with \textit{physics-guided architecture} and \textit{physics-guided aggregate supervision}. Cognizant of the cost of generation of drag force data, we aim to evaluate models in settings where there is a paucity of labelled training data.
\subsection{Physics-Guided Auxiliary Task Selection}\label{sec:multi_task_results}
When data about the target task is limited, we may employ exogenous inputs of processes that have an indirect influence over the target process to alleviate the effects of data paucity on model training. An effective way to achieve this is through multi-task learning. We first evaluate multi-task model performance relative to the corresponding single-task models to demonstrate performance gains.
\begin{table}[htpb]
    \small
    \begin{tabular}{|p{2.243cm}|c|c|p{2.9cm}|}
    \hline
         \textbf{Model}&\textbf{MSE}&\textbf{MRE}&\textbf{AU-REC}(\textbf{\% Imp.})\\ \hline
         Linear Reg.& 47.47& 38.48& 0.71332 (-19.54)\\\hline
         RF Reg.& 29.33& 19.13&0.82148 (-7.3)\\\hline
         GB Reg. & 25.02& 17.55&0.83692 (-5.60)\\\hline
         DNN& 20.50& 16.72&0.84573 (-4.61)\\\hline
         DNN-MT-Pres&20.12&15.66&0.85593 (-3.45)\\\hline
         DNN-MT-Vel&19.98&15.69&0.85556 (-3.49)\\\hline
        \ourmethodPXTX& 15.54& 14.06& 0.87232 (-1.61)\\\hline
         \ourmethodAll& \textbf{14.28}& \textbf{12.59}& \textbf{0.88657} (--)\\\hline
    \end{tabular}
    \caption{We compare the performance of \ourmethod and its variant \ourmethodPXTX (only x-components of pressure and shear drag are modeled) with many state-of-the-art regression baselines and show that the \ourmethod model yields significant performance improvement over all other models for the particle drag force prediction task. \iffalse We evaluate model performance in the context of three specific metrics described in Section~\ref{sec:experimental_setup}.\fi \iffalse We notice that \ourmethodAll model yields a significant performance improvement.\fi The last column of the table reports the AU-REC metric and also quantifies the percentage improvement of the best performing model i.e \ourmethod w.r.t all other models in the context of the AU-REC metric.}
    \label{tab:result_summary}
\end{table}
Table~\ref{tab:result_summary} shows the results of several multi-task and single task architectures that we tested to establish the superiority of multi-task models in the context of the particle drag force prediction task. It is widely known and accepted in physics that the drag force on each particle in fluid-particle systems such as the one being considered in this paper, is influenced strongly by the pressure and velocity fields acting on the particles~\cite{he2019supervised}. Hecnce, we wish to explicitly model the pressure and velocity fields around a particle, in addition to the main problem of predicting its drag force. To this end, we design two multi-task models, DNN-MT-Pres, DNN-MT-Vel, as described in section~\ref{sec:experimental_setup}. We notice that the two multi-task models DNN-MT-Pres and DNN-MT-Vel outperform not only the DNN model but also their single task counterparts (DNN+ Pres , DNN+ Vel). This improvement in performance may be attribured to the carefully selected auxiliary tasks to aid in learning the representation of the main task. This \textit{physics-guided auxiliary task selection} is also imperativie to our process of development of \ourmethod models to be detailed in section~\ref{sec:domain_informed_multitask_structural_priors}.
However, Table~\ref{tab:result_summary} also uncovers another interesting property which is the DNN+ models underperforming compared to their DNN-MT counterparts. Although the DNN+ and the DNN-MT models are predicting the same set of 11 values i.e 1 drag force value and 10 pressure or velocity samples in the vicinity of the particle, the DNN+ models make their predictions as part of a single task. Hence, the importance of the main task is diminished by the 10 additional auxiliary task values as the model tries to learn a jointly optimal representation. However, in the case of the DNN-MT models, each task has a set of disjoint hidden layers geared specifically towards learning the representation of the main task and a similar set of layers for the auxiliary task (in addition to a set of shared layers), which yields more flexibility in learning representations specific to the main and auxiliary task as well as a shared common representation. In addition, it is straightforward to explicitly control the effect of auxiliary tasks on the overall learning process in a multi-task setting.
\subsection{Physics-Guided Learning Architecture}\label{sec:domain_informed_multitask_structural_priors}
Section~\ref{sec:multi_task_results} showcases the effectiveness of multi-task learning and of \emph{physics-guided auxiliary task selection} for learning improved representations of particle drag force.

\par\noindent We now delve deeper and inspect the effects of expanding the realm of auxiliary tasks. In addition to this, we also use our domain knowledge regarding the physics of entities affecting the drag force acting on each particle, to influence model architecture through \textit{physics-guided structural priors}. As mentioned in Section~\ref{sec:problem_formulation}, \ourmethod has four carefully and deliberately chosen auxiliary tasks (pressure field prediction, velocity field prediction, predicting the pressure component(s) of drag, predicting the shear components of drag) aiding the main task of particle drag force prediction. In addition to this, the auxiliary tasks are arranged in a sequential manner to incorporate physical inter-dependencies among them leading up to the main task of particle drag force prediction. The effect of this carefully chosen physics-guided architecture and auxiliary tasks can be observed in Table~\ref{tab:result_summary}. We now inspect the different facets of this physics-guided architecture of the  \ourmethodAll model\footnote{\ourmethod was found to be robust to changes in auxiliary task hyperparameters, results included in appendix.}.
\begin{figure}[htpb]
    \centering
    \includegraphics[scale=0.31]{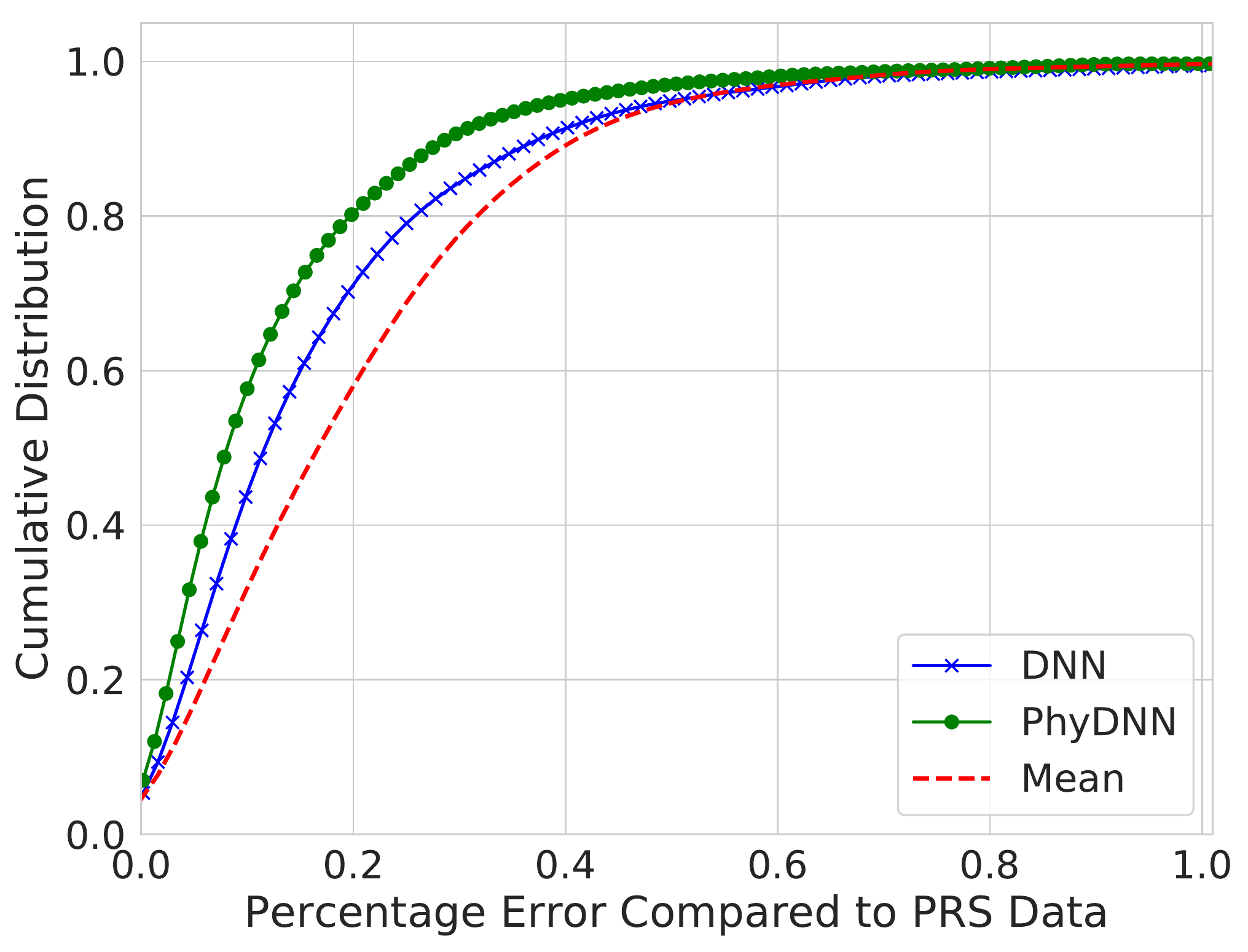}
    \caption{The cumulative distribution function of relative error for all (Re,$\,\phi$) combinations. \iffalse We notice that the Mean baseline (dotted red line) and the DNN model yield significantly higher relative error compared to the \ourmethod model.\fi  Overall, the \ourmethod model comfortably outperforms the DNN model and the Mean baseline (dotted red line).}
    \label{fig:rel_err_curve}
\end{figure}

\par\noindent We first characterize the performance of our \ourmethod models with respect to the DNN and mean baseline. Fig.~\ref{fig:rel_err_curve} represents the cumulative distribution of relative error of the predicted drag forces and the PRS ground truth drag force data. We notice that both DNN and \ourmethod outperform the mean baseline which essentially predicts the mean value per (Re,$\phi$) combination. The \ourmethod model significantly outperforms the DNN (current state-of-the-art~\cite{he2019supervised}) model to yield the best performance overall. We also tested DNN variants with dropout and $L_2$ regularization but found that performance deteriorated. Results were excluded due to space constraints.

\subsection{Performance With Limited Data}

Bearing in mind the high data generation cost of the PRS simulation, we wish to characterize an important facet of the \ourmethodAll model, namely, its ability to learn effective representations when faced with a paucity of training data. Hence, we evaluate the performance of the \ourmethodAll model as well as the other single task and multi-task DNN models, on different experimental settings obtained by continually reducing the fraction of data available for training the models. In  our  experiments,  the  training  fraction  was  reduced from  0.85  (i.e  85\%  of  the  data  used  for  training) to 0.35 (i.e  35\%  of  the  data  used  for  training).

\begin{figure}[!h]
    \centering
    \includegraphics[scale=0.24]{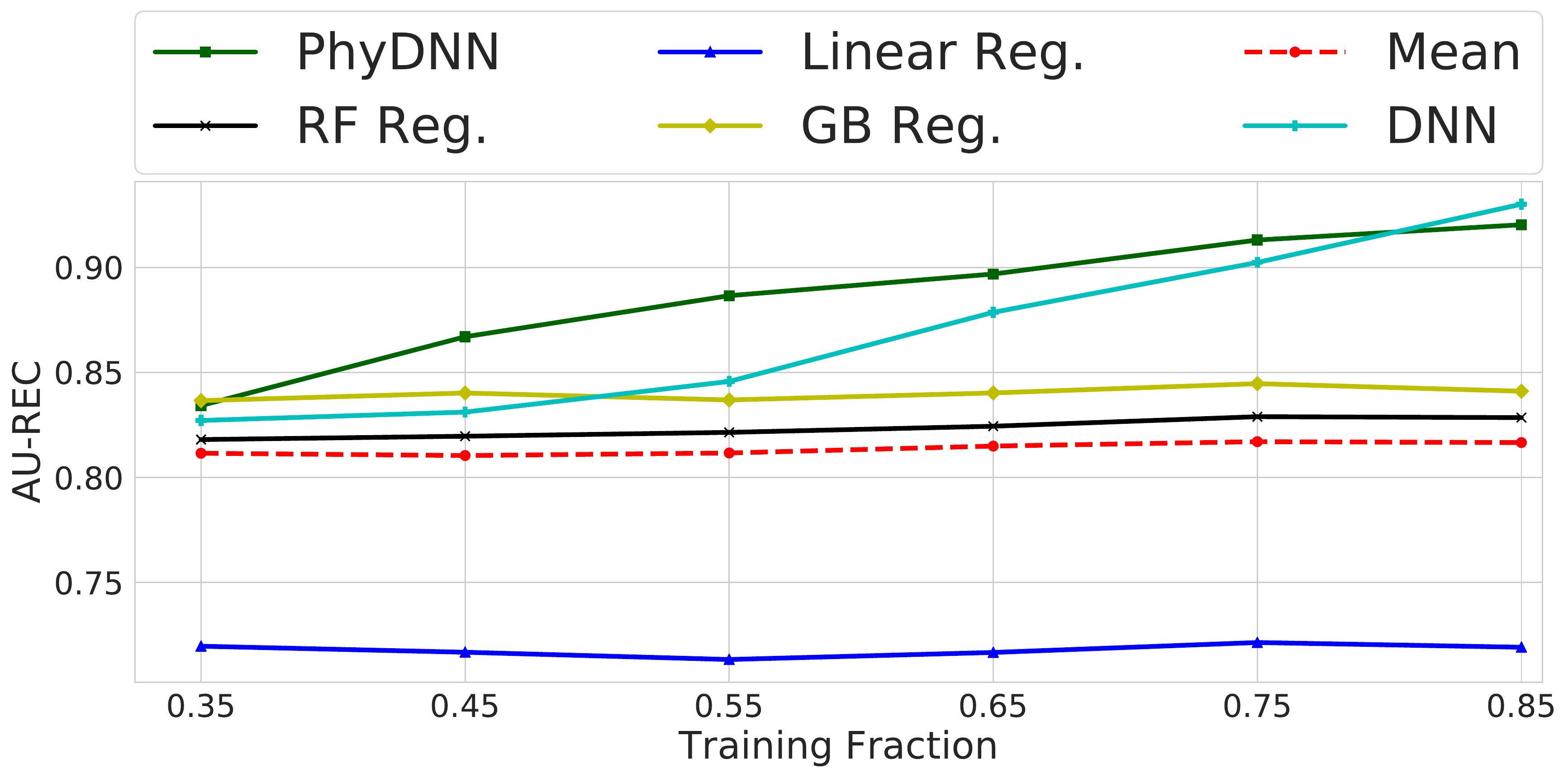}
    \caption{Model performance comparison for different levels of data paucity.}
    \label{fig:sparse_data_performance}
\end{figure}

\par\noindent Fig.~\ref{fig:sparse_data_performance} showcases the model performance in settings with limited data. We see that \ourmethodAll model significantly outperforms all other models in most settings (sparse and dense). We note that GB Reg. yields comparable performance to the \ourmethodAll model for the case of 0.35 training fraction. However, the gradient boosting (and all the other regression models except DNN) fail to learn useful information as more data is provided for training. We also notice that the DNN model fails to outperform the \ourmethodAll model for all but the last setting i.e the setting with 0.85 training fraction.   

\subsection{Characterizing \ourmethodAll Performance For Different (Re,$\,\phi$) Settings.}
In addition to quantitative evaluation, qualitative inspection is necessary for a deeper, holistic understanding of model behavior. Hence, we showcase the particle drag force predictions by the \ourmethodAll model for different (Re,$\phi$) combinations in Fig.~\ref{fig:prs_scatter_plots}\footnote{Figures for all (Re,$\phi$) combinations are in the appendix.}. We notice that the \ourmethodAll model yields accurate predictions (i.e yellow and red curves are aligned). This indicates that the \ourmethodAll model is able to effectively capture sophisticated particle interactions and the consequent effect of said interactions on the drag forces of the interacting particles. We notice that for high (Re,$\phi$) as in Fig.~\ref{fig:re_200_sf_35}, the drag force i.e PRS curve (yellow) is nonlinear in nature and that the magnitude of drag forces is also higher at higher (Re,$\phi$) settings. Such differing scales of drag force values can also complicate the drag force prediction problem as it is non-trivial for a single model to effectively learn such multi-modal target distributions. However, we find that the \ourmethodAll model is effective in this setting.  
\begin{figure}[htpb]
     \centering
     \begin{subfigure}[b]{0.48\columnwidth}
         \includegraphics[width=\textwidth]{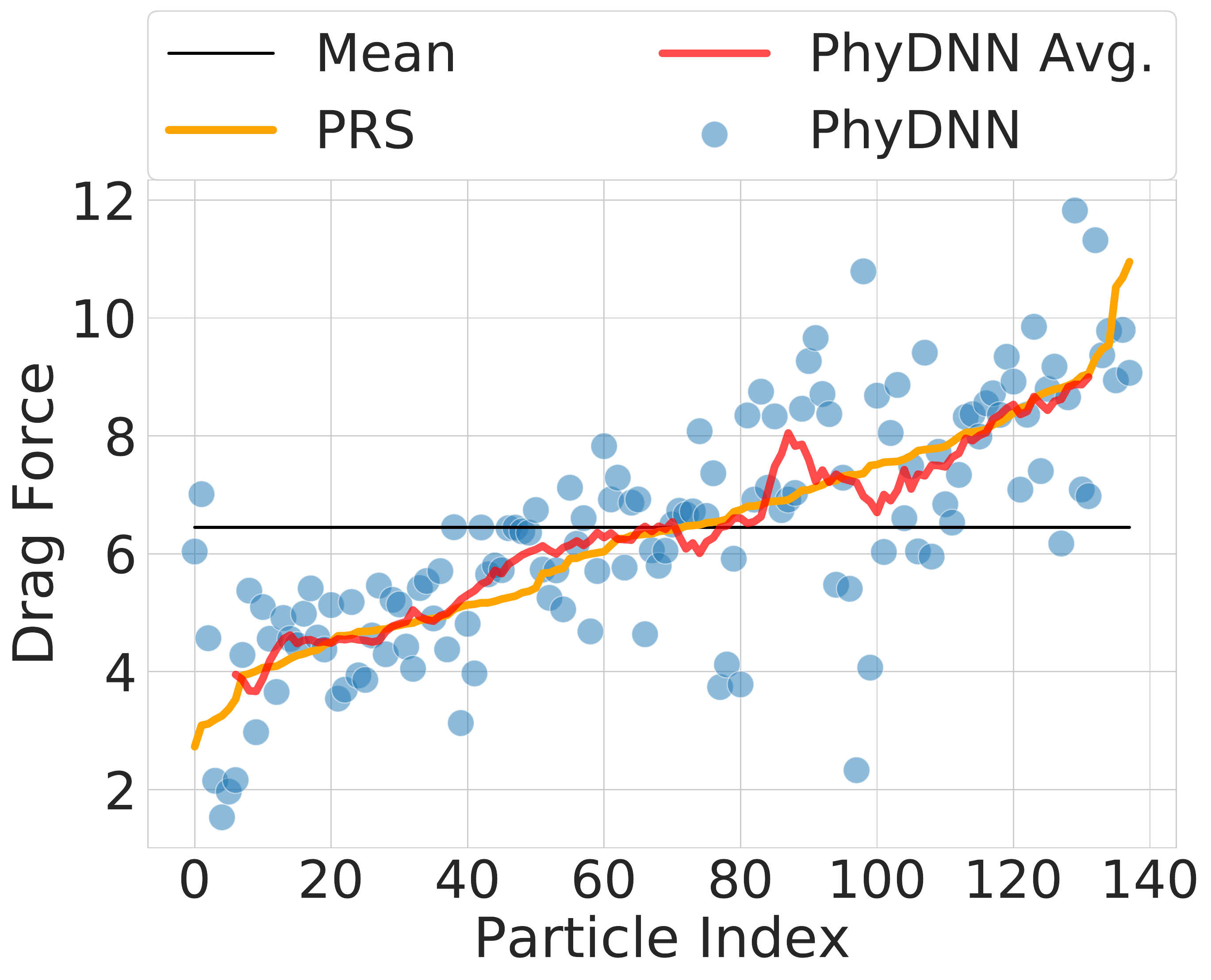}
         \caption{$Re = 10, \phi = 0.2$}
         \label{fig:re_10_sf_20}
     \end{subfigure}
     \begin{subfigure}[b]{0.48\columnwidth}
         \centering
         \includegraphics[width=\textwidth]{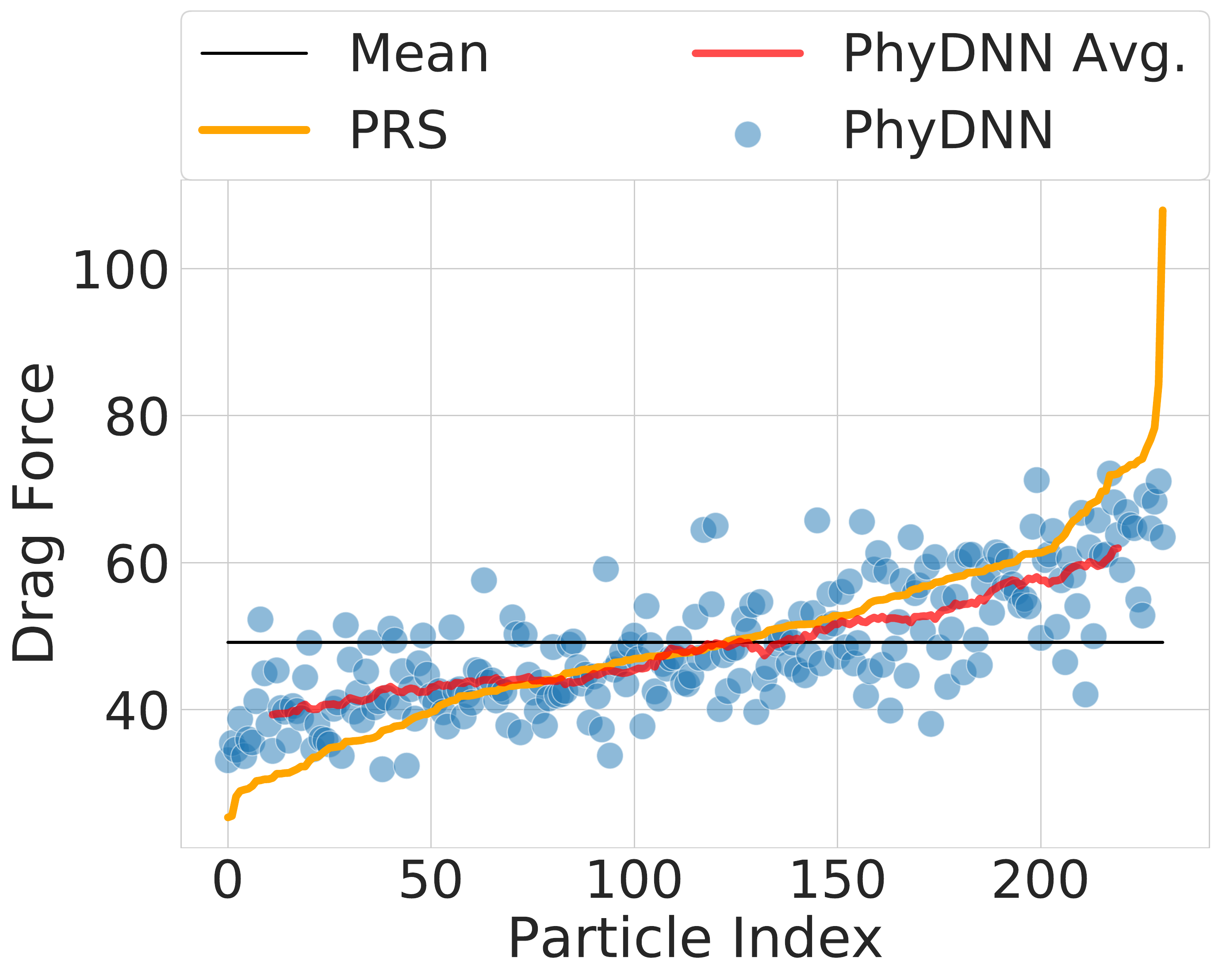}
         \caption{$Re = 200, \phi = 0.35$}
         \label{fig:re_200_sf_35}
     \end{subfigure}
        \caption{Each figure shows a comparison between \ourmethodAll predictions (red curve) and ground truth drag force data (yellow curve), for different (Re,$\phi$) cases. We also showcase the mean drag force value for each (Re,$\phi$) case (black). \iffalse The top row of figures indicates experiments conducted with low Re i.e Re=10 and different $\phi$ values. Notice that as $\phi$ increases, the number of samples is higher and hence the model is able to achieve a better representation of the corresponding PRS data curve (yellow).\fi  \iffalse We also notice that as Re and $\phi$ increase, the degree of non-linearity of the system increases due to the increase in complexity of the interactions between the particles. The magnitude of drag forces is also higher at higher Re and $\phi$ values.\fi}
        \label{fig:prs_scatter_plots}
\end{figure}
\begin{figure*}[htpb]
    \begin{subfigure}[b]{0.32\textwidth}
    \centering
    \includegraphics[width=0.95\textwidth]{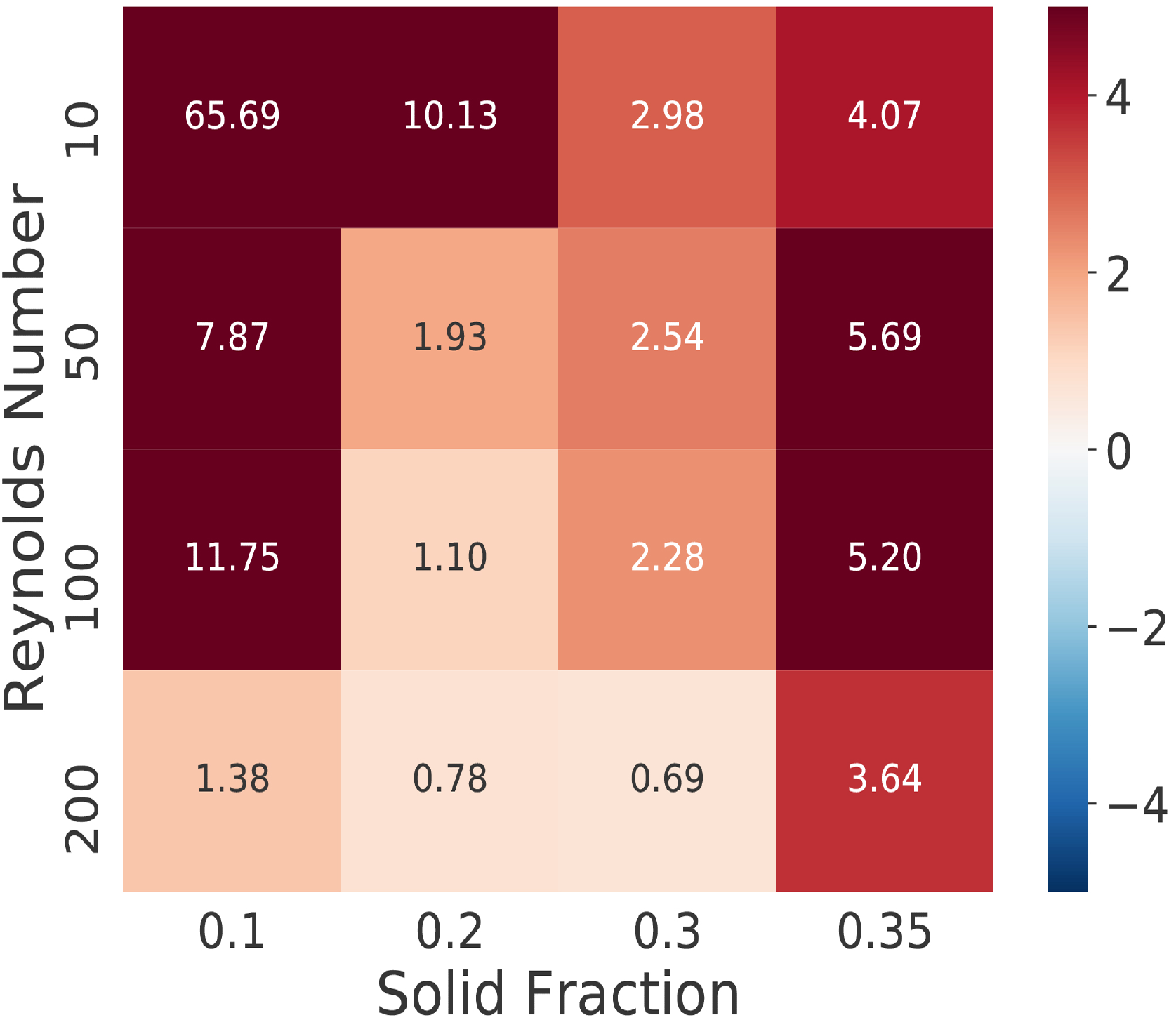}
    \caption{\ourmethodAll vs. DNN}
    \label{fig:aurec_ratio_comparison_phydnnall_vs_dnn}
    \end{subfigure}
    \hfill 
    \begin{subfigure}[b]{0.32\textwidth}
    \centering
    \includegraphics[width=0.95\textwidth]{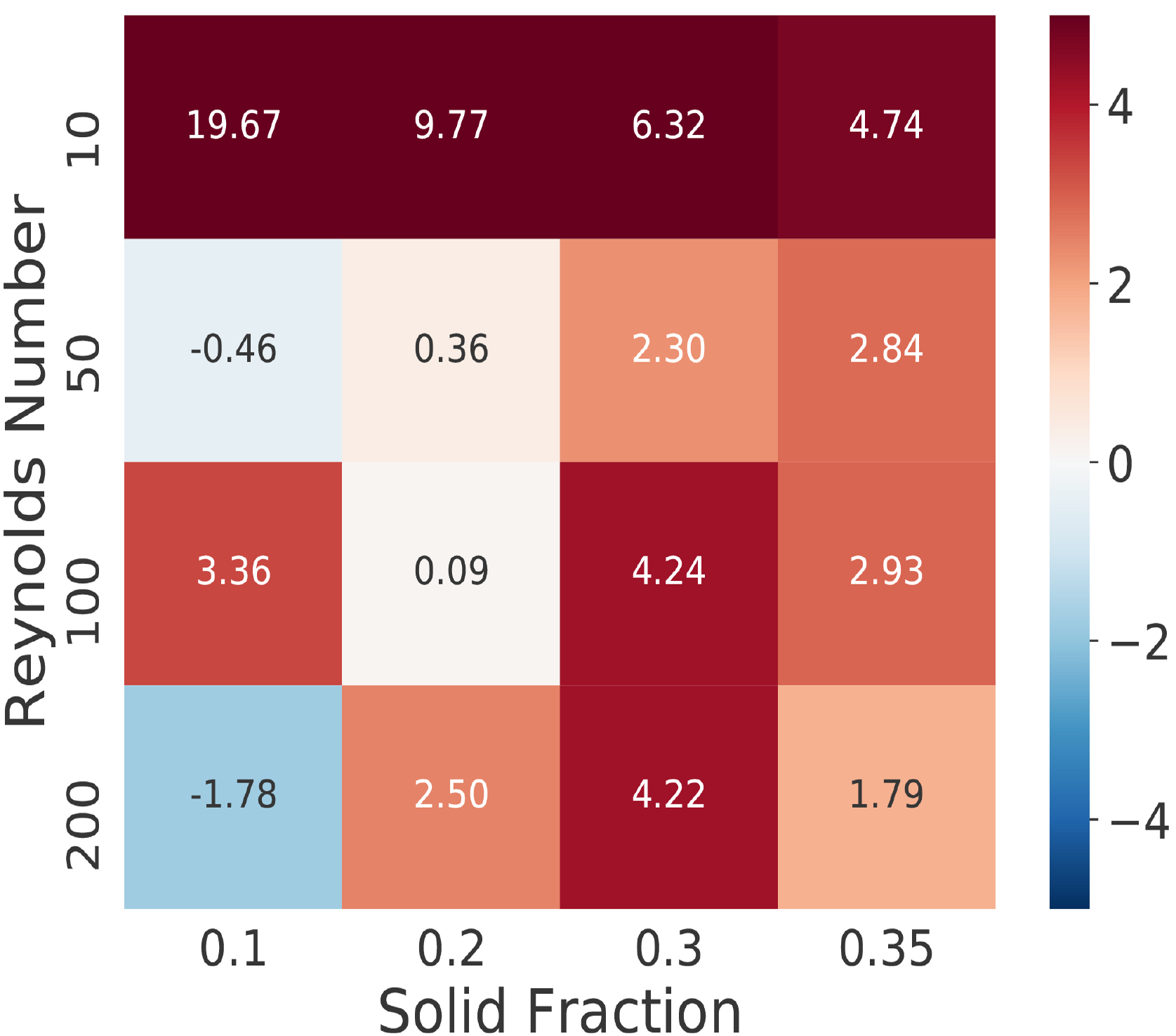}
    \caption{\ourmethodAll vs. DNN-MT-Pres.}
    \label{fig:aurec_raito_comparison_phydnnall_dnnmtpres}
    \end{subfigure}
    \hfill 
    \begin{subfigure}[b]{0.32\textwidth}
    \centering
    \includegraphics[width=0.95\textwidth]{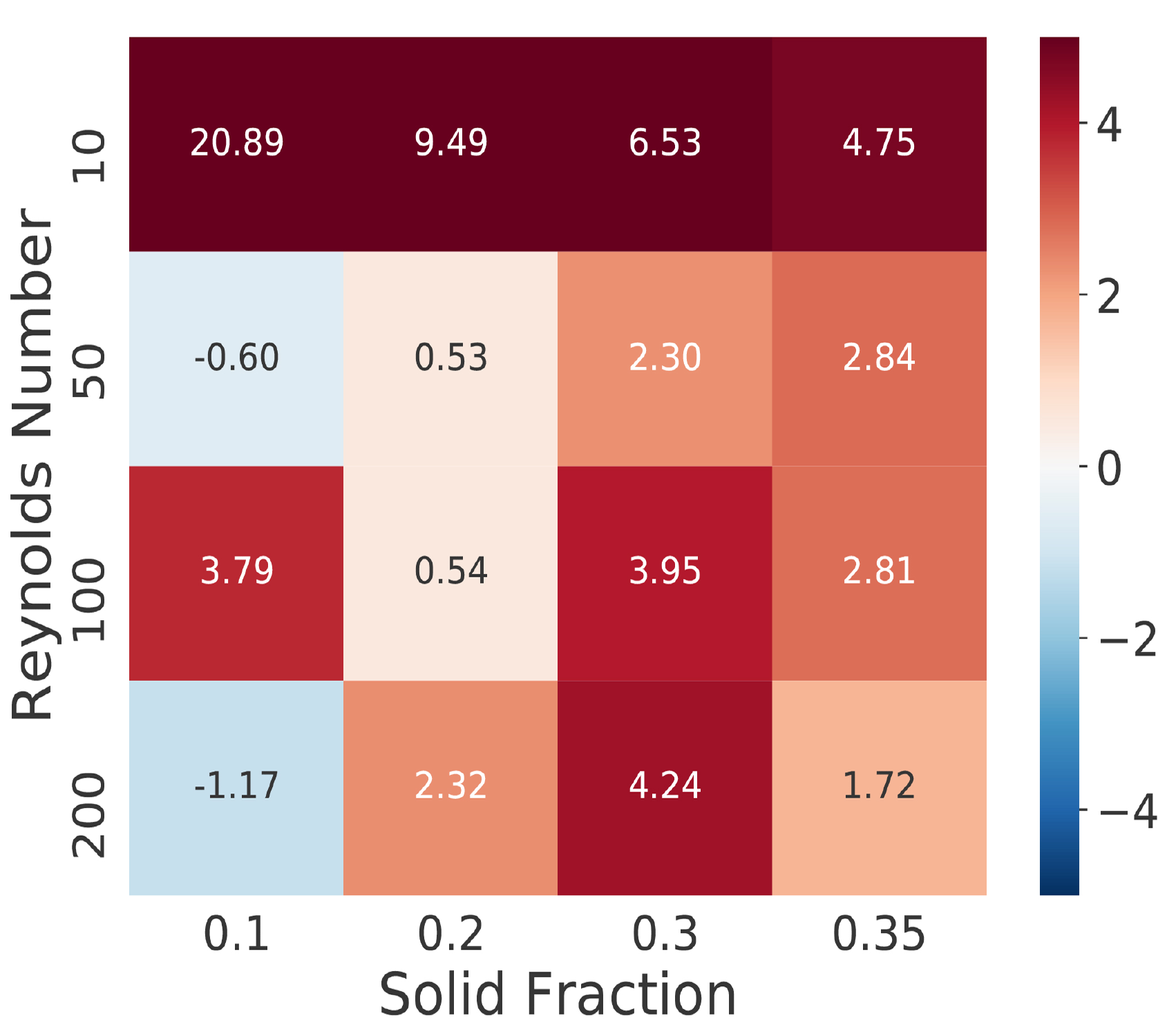}
    \caption{\ourmethodAll vs. DNN-MT-Vel}
    \label{fig:aurec_ratio_comparison_phydnnall_dnnmtvel}
    \end{subfigure}
    \caption{Each figure indicates the percentage improvement in the context of the AU-REC metric of the \ourmethodAll model over the DNN (Fig.~\ref{fig:aurec_ratio_comparison_phydnnall_vs_dnn}), DNN-MT-Pres (Fig.~\ref{fig:aurec_raito_comparison_phydnnall_dnnmtpres}) and DNN-MT-Vel (Fig.~\ref{fig:aurec_ratio_comparison_phydnnall_dnnmtvel}). Red squares show that \ourmethodAll does better and blue squares indicate that other models outperform \ourmethodAll. \ourmethodAll yields significant performance improvement over other models. \iffalse Specifically for low Reynolds numbers and high solid fractions (i.e higher number of particles), we notice significant performance improvement of \ourmethodAll.\fi  \iffalse  model is able to outperform other models indicating a better representative capacity of the complex particle interactions and the influences on particle drag therein. The DNN-MT-Pres and DNN-MT-Vel models are able to model the simple scenarios with fewer particles (i.e solid fraction = 0.1) well but underperform as the solidfraction (and hence the complexity of particle interactions) increases.\fi }
    \label{fig:aurec_ratio_comparison}
\end{figure*}

\par\noindent Thus far, we characterized the performance of the \ourmethodAll model in isolation for different (Re,$\phi$) contexts. In order to gain a deeper understanding of the performance of \ourmethod models for different (Re,$\phi$) combinations, we show percentage improvement for the AUREC metric of \ourmethodAll model and three other models in Fig.~\ref{fig:aurec_ratio_comparison_phydnnall_vs_dnn} - Fig.~\ref{fig:aurec_ratio_comparison_phydnnall_dnnmtvel}. We choose DNN, DNN-MT-Pres, DNN-MT-Vel as these are the closest by design to \ourmethodAll among all the baselines we consider in this paper. In Fig.~\ref{fig:aurec_ratio_comparison}, we see that \ourmethodAll outperforms the other models in most of the (Re,$\phi$) settings. \ourmethodAll when compared with the DNN model achieves especially good performance for low solid fraction settings which may be attributed to the inability of the DNN model to learn effectively with low data volumes as lower solid fractions have fewer training instances. In the case of the DNN-MT models, the \ourmethodAll model achieves significant performance improvement for high solid fraction ($\phi$ = \{0.2, 0.3, 0.35\}) cases and also for Re = \{10, 200\}, indicating that \ourmethodAll is able to perform well in the most complicated scenarios (high Re, high $\phi$). \ourmethodAll is able to achieve superior performance in 14 out of the 16 (Re, $\phi$) settings across all three models.

\subsection{Verifying Consistency With Domain Knowledge}
A significant advantage of \textit{physics-guided multi-task structural priors} is the increased interpretability provided by the resulting architecture. Since each component of the \ourmethodAll model has been designed and included based on sound domain theory, we may employ this theoretical understanding to verify through experimentation that the resulting behavior of each auxiliary component is indeed consistent with known theory.  
\begin{figure}[htpb]
    \centering
    \includegraphics[width=\columnwidth]{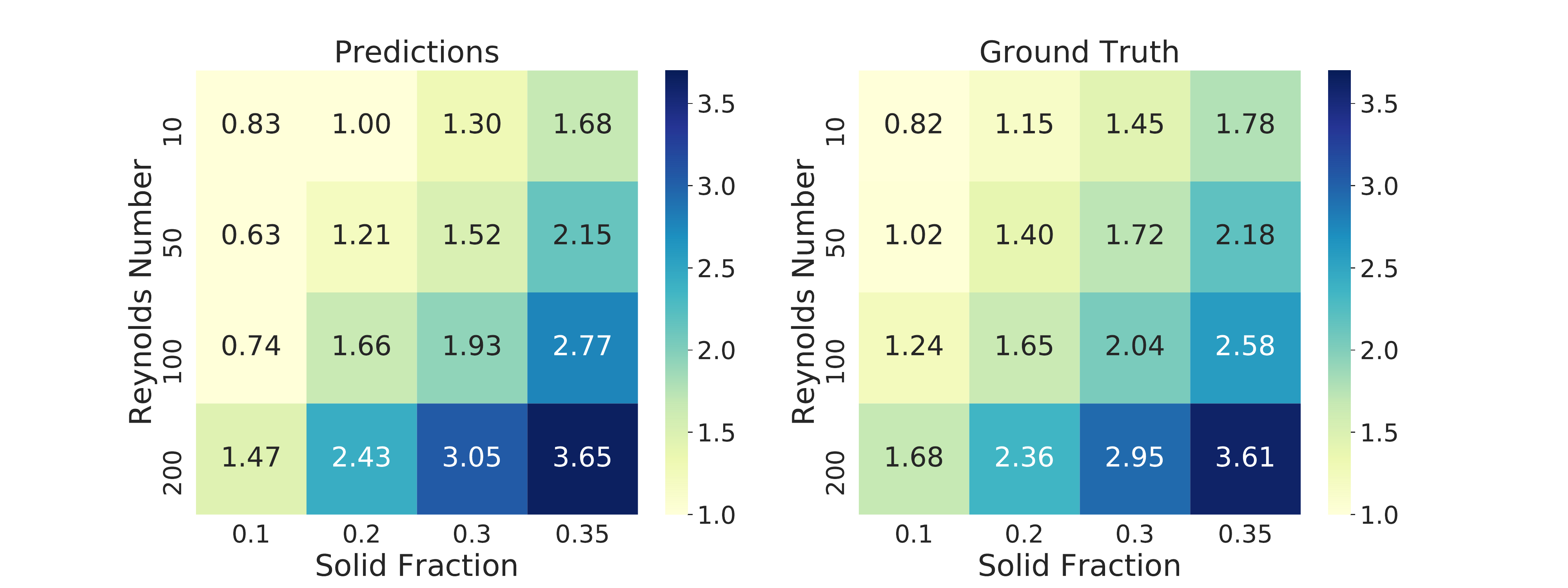}
    \caption{Heatmap showing ratio of absolute value of pressure drag ($F^P_x$) x-component to shear drag ($F^S_x$) x-component i.e \big($\frac{|F^P_x|}{|F^S_x|}$\big). Left figure shows ratio for \ourmethodAll predictions and the figure on the right shows the same ratio for ground truth data. We notice that distribution of ratios in both figures is almost identical.}
    \label{fig:heatmap_pressure_shear_component_ratio}
\end{figure}
We first verify the performance of the pressure and shear drag component prediction task in the \ourmethodAll model. It is well accepted in theory that for high Reynolds numbers, the proportion of the shear components of drag ($\mathbf{F^S}$) decreases~\cite{he2019supervised}. In order to evaluate this, we consider the ratio of the magnitude of the predicted pressure components in the x-direction ($F^P_x \in \mathbf{F^P}$) to the magnitude of the predicted shear components in the x-direction ($F^S_x \in \mathbf{F^S}$) for every (Re, $\phi$) setting\footnote{Similar behavior was recorded even when ratios were taken for all three pressure and shear drag components.}. The heatmap in Fig.~\ref{fig:heatmap_pressure_shear_component_ratio} depicts the comparison of this ratio of predicted pressure components to predicted shear components to a similar ratio derived from the ground truth pressure and shear components. We notice that there is good agreement between the predicted and ground truth ratios for each (Re, $\phi$) setting and also that the behavior of the predicted setting is indeed consistent with known domain theory as there is a noticeable decrease in the contribution of the shear components as we move toward high Re and high solid fraction $\phi$ settings.
\subsection{Auxiliary Representation Learning With Physics-Guided Statistical Constraints}
Two of the auxiliary prediction tasks involve predicting the pressure and velocity field samples around each particle. We hypothesized that since the drag force of a particle is influenced by the pressure and velocity fields, modeling them explicitly should help the model learn an improved representation of the main task of particle drag force prediction. 
\begin{figure*}[ht]
    \vspace{-0.25cm}
    \centering
    \includegraphics[width=\textwidth]{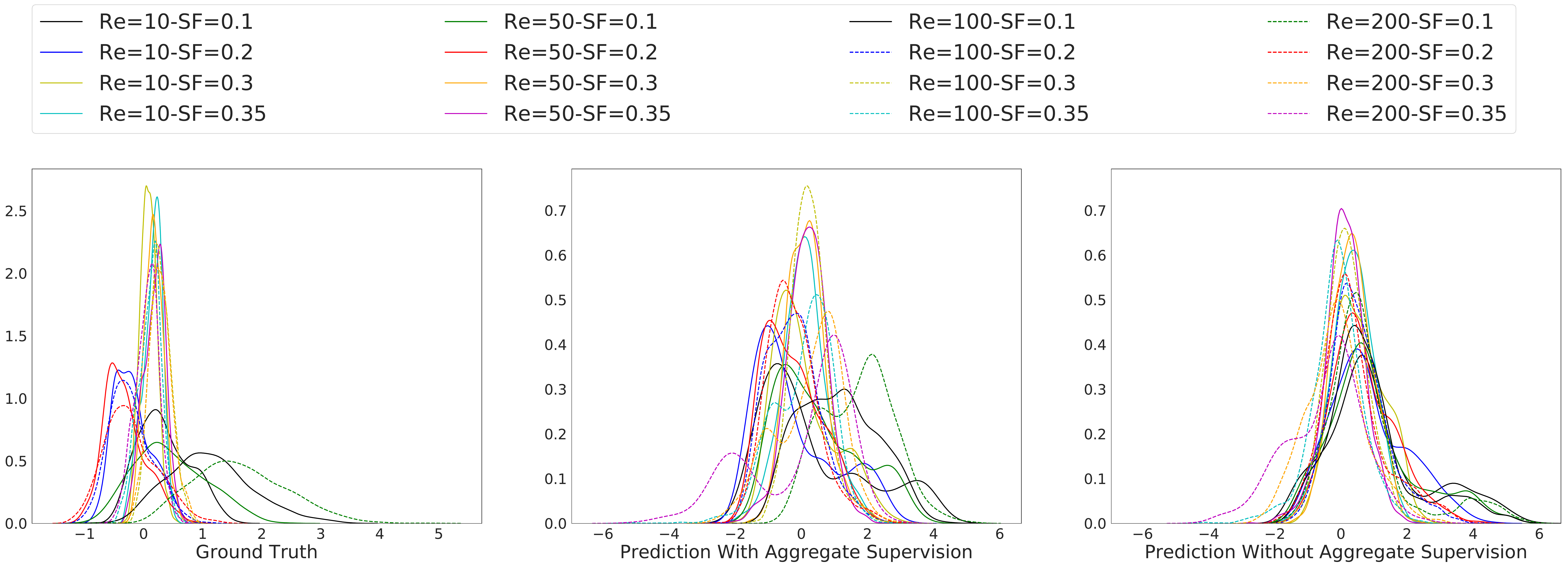}
    \caption{The figure depicts the densities of the ground truth (left) and predicted (center, right) pressure fields of the \ourmethodAll model for each (Re,$\phi$). Specifically, we wish to highlight the effect of aggregate supervision (physics-guided statistical prior) on the predicted pressure field. Notice that the PDFs of the pressure fields predicted with aggregate supervision are relatively more distributed similar to the ground truth distribution of pressure field PDFs as opposed to the plot on the right which represents predicted pressure field PDFs in the abscence of aggregate supervision and incorrectly depicts a some what uniform behavior for all the PDFs of different (Re,$\phi$) cases.}
    \label{fig:pressure_field_pdf}
\end{figure*}
In Fig.~\ref{fig:pressure_field_pdf}, we notice that ground-truth pressure field PDFs exhibit a grouped structure. Interestingly, the pressure field PDFs can be divided into three distinct groups with all the pressure fields with $\phi=0.2$ being grouped to the left of the plot, pressure fields with $\phi=0.1$ being grouped toward the bottom, right of the plot and the rest of the PDFs forming a core (highly dense) group in the center. Hence, we infer that solid fraction has a significant influence on the pressure field. It is non-trivial for models to automatically replicate such multi-modal and grouped behavior and hence we introduce \textit{physics-guided statistical priors} through aggregate supervision during model training of \ourmethodAll. We notice that the learned distribution with aggregate supervision Fig.~\ref{fig:pressure_field_pdf} (center) has a similar grouped structure to the ground truth PDF pressure field. For the purpose of comparison, we also obtained the predicted pressure field PDFs of a version of \ourmethodAll trained without aggregate supervision and the result is depicted in Fig.~\ref{fig:pressure_field_pdf} (right). We notice that the PDFs exhibit a kind of \textit{mode collapse} behavior and do not display any similarities to ground truth pressure field PDFs. Similar aggregate supervision was also applied to the velocity field prediction task and we found that incorporating physics-guided aggregate supervision to ensure learning representations consistent with theory, led to significantly improved model performance.

%% file: sections/conclusion.tex
In this paper, we introduce \ourmethodAll. A physics inspired deep learning model developed to incorporate fluid mechanical theory into the model architecture and proposed physics informed auxiliary tasks selection to aid with training under data paucity. We conduct a rigorous analysis to test \ourmethodAll performance in settings with limited labelled data and find that \ourmethodAll significantly outperforms all other state-of-the-art regression baselines for the task of particle drag force prediction achieving an average performance improvement of \textbf{8.46\%} across all models. We verify that each physics informed auxiliary task of \ourmethodAll is consistent with existing physics theory, yielding greater model interpretability. Finally, we also showcase the effect of augmenting \ourmethodAll with physics-guided aggregate supervision to constrain auxiliary tasks to be consistent with ground truth data. In future, we will augment \ourmethodAll with more sophisticated learning architectures.  

%% file: main.bbl
\begin{thebibliography}{10}
\providecommand{\url}[1]{#1}
\csname url@samestyle\endcsname
\providecommand{\newblock}{\relax}
\providecommand{\bibinfo}[2]{#2}
\providecommand{\BIBentrySTDinterwordspacing}{\spaceskip=0pt\relax}
\providecommand{\BIBentryALTinterwordstretchfactor}{4}
\providecommand{\BIBentryALTinterwordspacing}{\spaceskip=\fontdimen2\font plus
\BIBentryALTinterwordstretchfactor\fontdimen3\font minus
  \fontdimen4\font\relax}
\providecommand{\BIBforeignlanguage}[2]{{%
\expandafter\ifx\csname l@#1\endcsname\relax
\typeout{** WARNING: IEEEtran.bst: No hyphenation pattern has been}%
\typeout{** loaded for the language `#1'. Using the pattern for}%
\typeout{** the default language instead.}%
\else
\language=\csname l@#1\endcsname
\fi
#2}}
\providecommand{\BIBdecl}{\relax}
\BIBdecl

\bibitem{he2017evaluation}
L.~He, D.~K. Tafti, and K.~Nagendra, ``Evaluation of drag correlations using
  particle resolved simulations of spheres and ellipsoids in assembly,''
  \emph{Powder technology}, vol. 313, 2017.

\bibitem{he2019supervised}
L.~He and D.~K. Tafti, ``A supervised machine learning approach for predicting
  variable drag forces on spherical particles in suspension,'' \emph{Powder
  technology}, vol. 345, 2019.

\bibitem{li2003gas}
J.~Li and J.~Kuipers, ``Gas-particle interactions in dense gas-fluidized
  beds,'' \emph{Chemical Engineering Science}, vol.~58, no. 3-6, 2003.

\bibitem{wen1966yh}
C.~Wen, ``Yh yu. mechanics of fluidization,'' in \emph{Chemical Engineering
  Progress Symposium Series}, vol.~62, no.~62, 1966.

\bibitem{di1994voidage}
R.~Di~Felice, ``The voidage function for fluid-particle interaction systems,''
  \emph{International Journal of Multiphase Flow}, vol.~20, no.~1, 1994.

\bibitem{tenneti2011drag}
S.~Tenneti, R.~Garg, and S.~Subramaniam, ``Drag law for monodisperse gas--solid
  systems using particle-resolved direct numerical simulation of flow past
  fixed assemblies of spheres,'' \emph{International journal of multiphase
  flow}, vol.~37, no.~9, 2011.

\bibitem{wong2009active}
K.~C. Wong, L.~Wang, and P.~Shi, ``Active model with orthotropic hyperelastic
  material for cardiac image analysis,'' in \emph{Functional Imaging and
  Modeling of the Heart}.\hskip 1em plus 0.5em minus 0.4em\relax Springer,
  2009.

\bibitem{xu2015robust}
J.~Xu \emph{et~al.}, ``Robust transmural electrophysiological imaging:
  Integrating sparse and dynamic physiological models into ecg-based
  inference,'' in \emph{MICCAI}.\hskip 1em plus 0.5em minus 0.4em\relax
  Springer, 2015.

\bibitem{denli2014multi}
H.~Denli \emph{et~al.}, ``Multi-scale graphical models for spatio-temporal
  processes,'' in \emph{NeurIPS}, 2014.

\bibitem{chatterjee2012sparse}
S.~Chatterjee \emph{et~al.}, ``Sparse group lasso: Consistency and climate
  applications.'' in \emph{SDM12'}.\hskip 1em plus 0.5em minus 0.4em\relax
  SIAM, 2012.

\bibitem{liu2013accounting}
J.~Liu \emph{et~al.}, ``Accounting for linkage disequilibrium in genome-wide
  association studies: a penalized regression method,'' \emph{Statistics and
  its interface}, vol.~6, no.~1, 2013.

\bibitem{majda2012physics}
A.~J. Majda and J.~Harlim, ``Physics constrained nonlinear regression models
  for time series,'' \emph{Nonlinearity}, vol.~26, no.~1, 2012.

\bibitem{majda2012fundamental}
A.~J. Majda and Y.~Yuan, ``Fundamental limitations of ad hoc linear and
  quadratic multi-level regression models for physical systems,''
  \emph{Discrete and Continuous Dynamical Systems B}, vol.~17, no.~4, 2012.

\bibitem{waterman1986guide}
D.~Waterman, ``A guide to expert systems,'' 1986.

\bibitem{abu1990learning}
Y.~S. Abu-Mostafa, ``Learning from hints in neural networks,'' \emph{Journal of
  complexity}, vol.~6, no.~2, 1990.

\bibitem{chen2018neural}
T.~Q. Chen \emph{et~al.}, ``Neural ordinary differential equations,'' in
  \emph{NeurIPS}, 2018.

\bibitem{zhu2018convolutional}
M.~Zhu, B.~Chang, and C.~Fu, ``Convolutional neural networks combined with
  runge-kutta methods,'' \emph{arXiv:1802.08831}, 2018.

\bibitem{karpatne2017theory}
A.~Karpatne \emph{et~al.}, ``Theory-guided data science: A new paradigm for
  scientific discovery from data,'' \emph{IEEE TKDE}, vol.~29, no.~10, 2017.

\bibitem{Ren_Stewart_Song_Kuleshov_Ermon_2018}
H.~Ren \emph{et~al.}, ``Learning with weak supervision from physics and
  data-driven constraints.'' \emph{AI Magazine}, vol.~39, no.~1, 2018.

\bibitem{stewart2017label}
R.~Stewart and S.~Ermon, ``Label-free supervision of neural networks with
  physics and domain knowledge,'' in \emph{AAAI}, 2017.

\bibitem{ioannou2018structural}
Y.~A. Ioannou, ``Structural priors in deep neural networks,'' Ph.D.
  dissertation, University of Cambridge, 2018.

\bibitem{ling2016reynolds}
J.~Ling, A.~Kurzawski, and J.~Templeton, ``Reynolds averaged turbulence
  modelling using deep neural networks with embedded invariance,''
  \emph{Journal of Fluid Mechanics}, vol. 807, 2016.

\bibitem{seo2019differentiable}
S.~Seo and Y.~Liu, ``Differentiable physics-informed graph networks,''
  \emph{arXiv:1902.02950}, 2019.

\bibitem{leibo2017view}
J.~Z. Leibo \emph{et~al.}, ``View-tolerant face recognition and hebbian
  learning imply mirror-symmetric neural tuning to head orientation,''
  \emph{Current Biology}, vol.~27, no.~1, 2017.

\bibitem{anderson2019cormorant}
B.~Anderson, T.-S. Hy, and R.~Kondor, ``Cormorant: Covariant molecular neural
  networks,'' \emph{arXiv:1906.04015}, 2019.

\bibitem{kondor2018generalization}
R.~Kondor and S.~Trivedi, ``On the generalization of equivariance and
  convolution in neural networks to the action of compact groups,''
  \emph{arXiv:1802.03690}, 2018.

\bibitem{scikitlearn}
F.~Pedregosa \emph{et~al.}, ``Scikit-learn: Machine learning in {P}ython,''
  \emph{JMLR}, vol.~12, 2011.

\end{thebibliography}
